%% file: DisFace3DNet.tex
\documentclass[journal,10pt,letterpaper]{IEEEtran}
\input{disface_preamble.tex}
\hypersetup{
  hidelinks,
  pdftitle={DisFace3DNet: Explainable Facial Attractiveness Prediction via 3D Component Disentanglement},
  pdfauthor={Fenggui Rao, Yan Luximon, and Jie Zhang},
  pdfsubject={Explainable facial attractiveness prediction via 3D component disentanglement},
  pdfkeywords={facial attractiveness explanation, facial attractiveness prediction, 3D component disentanglement}
}

\InputIfFileExists{supplement_refs.tex}{}{}
\begin{document}
\bstctlcite{disfaceIEEEcontrol}
\raggedbottom
% Use the standard hyphenated-line break penalty; a hard ban leaves column gaps.
\brokenpenalty=100

\title{\DisFacePaperTitle}
\author{Fenggui Rao~\orcidlink{0009-0000-8990-1515},
Yan Luximon~\orcidlink{0000-0003-2843-847X},
and Jie Zhang~\orcidlink{0000-0001-8219-5590},~\IEEEmembership{Member,~IEEE}
\thanks{This work was supported by the grants from the Research Grants Council of the Hong Kong Special Administrative Region (Project No. GRF/PolyU 15616124), the Science and Technology Development Fund of Macau SAR (Project No. 0057/2025/ITP1) and Macao Polytechnic University (Project No. RP/FCA-15/2025).}
\thanks{The human evaluation was approved by The Hong Kong Polytechnic University Institutional Review Board (IRB) through the Human Subjects Ethics Application and Review System (HSEARS; Protocol No. HSEARS20240313013).}
\thanks{Fenggui Rao is with the School of Design, The Hong Kong Polytechnic University, Hong Kong, China (e-mail: \mbox{robbie.rao@connect.polyu.hk}).}
\thanks{Yan Luximon is with the School of Design, The Hong Kong Polytechnic University, Hong Kong, China, and the Laboratory for Artificial Intelligence in Design, Hong Kong Science Park, Hong Kong, China (e-mail: yan.luximon@polyu.edu.hk).}
\thanks{Jie Zhang is with the Faculty of Applied Sciences, Macao Polytechnic University, Macau, China (e-mail: jpeter.zhang@mpu.edu.mo).}
\thanks{Corresponding authors: Yan Luximon and Jie Zhang.}}

\markboth{IEEE Transactions on Image Processing}%
{Rao \MakeLowercase{\textit{et al.}}: DisFace3DNet}

\maketitle
\thispagestyle{empty}

\begin{abstract}
Facial attractiveness prediction usually assigns one overall rating, leaving the roles of shape, appearance, and viewing conditions implicit. We propose DisFace3DNet, which uses 3D component disentanglement to learn seven component reference scores from overall ratings with auxiliary weak semantic supervision, without human-labeled component targets. Designated 3D representations and image cues feed jointly learned routes for identity, skin, hair, light, background, expression, and pose. A constrained fit then combines five static and two signed dynamic scores into the overall rating, exposing each component's numerical contribution and supporting component-specific comparisons across images. On SCUT-FBP5500, DisFace3DNet achieves a Pearson correlation of $0.8904\pm0.0063$ (mean $\pm$ standard deviation across five folds) with average human ratings; its component terms reconstruct every held-out prediction to numerical precision. Skin, hair, and facial shape account for the largest component-wise prediction variation. Human evaluation supports the score directions for facial shape, skin, and hair; expression agreement is weaker. DisFace3DNet thus connects overall prediction to quantitative analysis of the facial and contextual cues entering each estimate.
\end{abstract}

\begin{IEEEkeywords}
Facial attractiveness explanation, facial attractiveness prediction, 3D component disentanglement
\end{IEEEkeywords}

\section{Introduction}\label{sec:introduction}
Facial attractiveness prediction (FAP) aims to estimate human attractiveness evaluations from face images~\cite{eisenthal2006facial,Liang2018SCUT}, while facial attractiveness explanation (FAE) seeks to identify the visual factors associated with these evaluations and to understand how predictive models use such information. Both tasks are relevant to the study and modeling of facial perception: attractiveness evaluations vary with facial shape and texture~\cite{nakamura2019datadriven}, expression, and pose~\cite{sutherland2017facial}. Digital-human design~\cite{wang2025mega}, hairstyle try-on~\cite{khwanmuang2023salon}, and live-video retouching~\cite{li2025facial} involve comparing alternative facial appearances. Prediction provides a quantitative basis for such comparisons, while explanation helps identify the visual features to consider when refining a design or retouching setting.

\begin{figure*}[t]
  \centering
  \captionsetup{skip=0pt}
  \includegraphics[width=\textwidth]{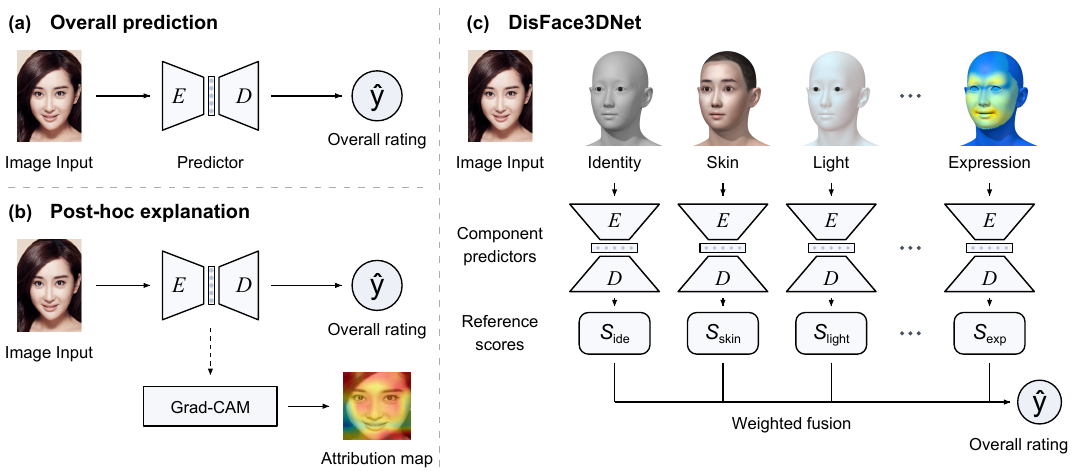}
  \caption{Prediction and explanation interfaces. (a) Overall prediction through regression, distribution learning, or ranking~\cite{eisenthal2006facial,fan2018label,lyu2024ranking}. (b) Post-hoc explanation~\cite{Selvaraju2017GradCAM,Chattopadhyay2018GradCAMPlusPlus,Wang2020ScoreCAM}, illustrated by Grad-CAM. (c) DisFace3DNet uses separate component predictors to produce component reference scores, which are combined by weighted fusion into the overall rating. Head renderings illustrate the components; scores are symbolic. $E$ and $D$ denote encoder and decoder.}
  \label{fig:prediction_explanation}
\end{figure*}

Existing FAP methods predict overall ratings through regression, retain rating spread through label-distribution learning, and model relative preferences through ranking~\cite{eisenthal2006facial,fan2018label,lin2019regression,lyu2024ranking}. Geometric, attribute-aware, and multimodal features enrich their representations~\cite{Lin2019AaNet,peng2024geometric,li2025facial}. However, their overall predictions do not reveal separately reportable scores for facial shape, appearance, and capture conditions, or how these scores contribute to the predicted rating.
Post-hoc explanation methods use visual attribution, concept analysis, or image editing to identify relevant regions, semantic sensitivities, or appearance changes~\cite{Selvaraju2017GradCAM,kim2018tcav,peng2024isfbgan}. Part-based prediction methods also learn regional sub-scores~\cite{boukhari2025scat}. Figure~\ref{fig:prediction_explanation} contrasts overall prediction (a), post-hoc explanation (b), and DisFace3DNet (c). Our objective is to separate 3D structure, surface appearance, and viewing conditions into component scores with explicit additive roles in prediction. Obtaining reliable component-level training labels, however, poses a fundamental challenge: facial shape, albedo, and lighting jointly determine a 2D face image, making it difficult for human annotators to assign an absolute attractiveness score to each factor in isolation. Attribute-focused pairwise comparisons can test relative preferences, but do not supply independently calibrated scores for these abstract components.

To address these limitations, we propose DisFace3DNet. To the best of our knowledge, DisFace3DNet is the first model to use 3D component disentanglement for explainable facial attractiveness prediction through jointly learned component scores and their explicit additive contributions. Designated 3D and image inputs separate structural, appearance, and viewing-condition evidence. Auxiliary weak semantic labels orient the component scores under overall-rating supervision, and constrained fusion fixes their numerical relationship to the overall prediction.
\par
This paper makes three contributions:
\begin{itemize}
\setlength{\itemsep}{0.15em}
\setlength{\parskip}{0pt}
  \item We propose DisFace3DNet\footnote{Project page: \url{https://github.com/RobbieRao/DisFace3DNet}.}, an explainable facial attractiveness prediction model based on 3D component disentanglement that produces seven separately reportable component reference scores, supporting component-specific image comparisons and analysis of each component's contribution to the overall rating.
  \item We develop designated 3D and image inputs with jointly learned branches for five static and two signed dynamic components. Auxiliary semantic guidance orients the scores, and constrained fusion yields explicit additive prediction terms.
  \item We evaluate prediction accuracy and use matched controls to test inputs, semantic guidance, and joint learning alongside component correspondence and dynamic-score variation. Human evaluation supports the score directions for identity, skin, and hair.
\end{itemize}

\section{Related Work}\label{sec:related_work}
\subsection{Facial Attractiveness Prediction}
Facial attractiveness prediction methods estimate overall human ratings, rating distributions, or relative preferences from face images~\cite{laurentini2014computer,liu2016advances,boukhari2025comprehensive}. Regression-based methods map handcrafted or learned facial features to mean ratings~\cite{eisenthal2006facial,wang2014attractive,xu2017facial}; label-distribution learning methods preserve rating spread~\cite{fan2018label,chen2019facial}, and ranking-based methods incorporate relative preferences~\cite{lin2019regression,lyu2024ranking}. All three method types learn from overall attractiveness evaluations.

Convolutional neural network (CNN) methods use searched architectures, network ensembles, or ensembles trained with different losses~\cite{zhang2022nas4fbp,bougourzi2022cnn,bougourzi2022dynamic,saeed2023automatic}. Transformer-based methods model image-wide dependencies~\cite{liu2023fbpformer}, with TransFBP adding cross-attention and attention-guided augmentation~\cite{boukhari2026enhancing}. Attention-based methods emphasize informative cues through co-attention~\cite{shi2019improving}, spatial/channel weighting~\cite{cao2020deep}, or dynamic attentive convolution~\cite{sun2024dynamic}. Other methods incorporate attributes through generated filters~\cite{Lin2019AaNet}, share features across tasks~\cite{xu2019hierarchical,vahdati2020facial}, or learn from paired images~\cite{li2024broad}, while retaining overall-score outputs. Geometry-aware methods use shape--texture fusion~\cite{xiao2021beauty3dfacenet,liu20243d}, structural measurements~\cite{liu2022computation}, or depth-aware landmarks~\cite{xie2025crucial}. GPNet combines full-face and local features with landmark-based regularization~\cite{peng2024geometric}. Multimodal methods such as FPEM fuse face-prior and vision--language features~\cite{li2025facial}.

These methods use geometric, attribute, and semantic cues for overall prediction, leaving factor-specific attractiveness scores and contributions implicit. DisFace3DNet addresses this gap through 3D component disentanglement and jointly learned component scoring branches. Designated inputs separate component evidence, while constrained fusion links the seven reference scores to the overall rating through explicit additive terms, enabling component-specific comparisons and contribution analysis.

\subsection{Facial Attractiveness Explanation}
FAE studies examine the factors underlying human evaluation and model predictions. Perceptual studies consider shape, averageness, and skin appearance~\cite{langlois2000maxims,rhodes2006evolutionary,little2011facial}, as well as expression, head tilt, imaging conditions, and observer context~\cite{marshall2020human,sutherland2017facial,sutherland2018facial,Sutherland2022Understanding}. Feature-level analyses examine shape categories, structural measurements, skin texture, and geometric morphometrics~\cite{zhao2020novel,sano2023computational}; an artificial-face trustworthiness review discusses related feature and evaluation-design effects~\cite{wu2026facial}. Controlled 3D studies vary shape and reflectance separately to measure perceptual responses~\cite{nakamura2019datadriven,oh2019contributions}. These studies motivate factor selection; explaining a model prediction also requires specifying each factor's role in the scoring rule.

Post-hoc explanation methods analyze a trained prediction model. Grad-CAM uses gradients to weight feature maps and localize relevant regions~\cite{Selvaraju2017GradCAM}; Grad-CAM++, Score-CAM, and Ablation-CAM use alternative localization mechanisms~\cite{Chattopadhyay2018GradCAMPlusPlus,Wang2020ScoreCAM,Desai2020AblationCAM}. Concept-based methods measure sensitivity along learned semantic directions~\cite{kim2018tcav}, while beautification and attractiveness-guided editing methods reveal score-associated appearance changes~\cite{peng2024isfbgan,li2025diffusionfae}. These post-hoc methods characterize model responses without assigning component attractiveness scores. Part-based explanation methods instead learn sub-scores within a prediction model. SCAT~\cite{boukhari2025scat} uses multilayer perceptrons (MLPs) for facial-part sub-scores and a Transformer to aggregate 2D part embeddings into an overall rating. A consistency loss encourages SCAT's overall rating to match the mean sub-score, but does not enforce an exact additive prediction rule. The facial-part decomposition also does not separate shape, appearance, and viewing conditions.

Concept bottleneck models connect supervised intermediate concepts to predictions~\cite{koh2020concept}. For FAE, the remaining gap is to connect separated factors to component scores and explicit contributions. DisFace3DNet builds designated inputs from reconstructed geometry, albedo, lighting, pose, and expression~\cite{bai2023ffhquv,deng2019deep3dface,wang2024geometric3ddfa}, together with segmented hair and background. Weak semantic guidance orients component scores under overall-rating supervision, and constrained fusion makes their additive roles explicit without human-labeled component targets.

\section{Method} \label{sec:method}

\subsection{Overview}
\begin{figure*}[!t]
  \centering
  \includegraphics[width=0.98\linewidth]{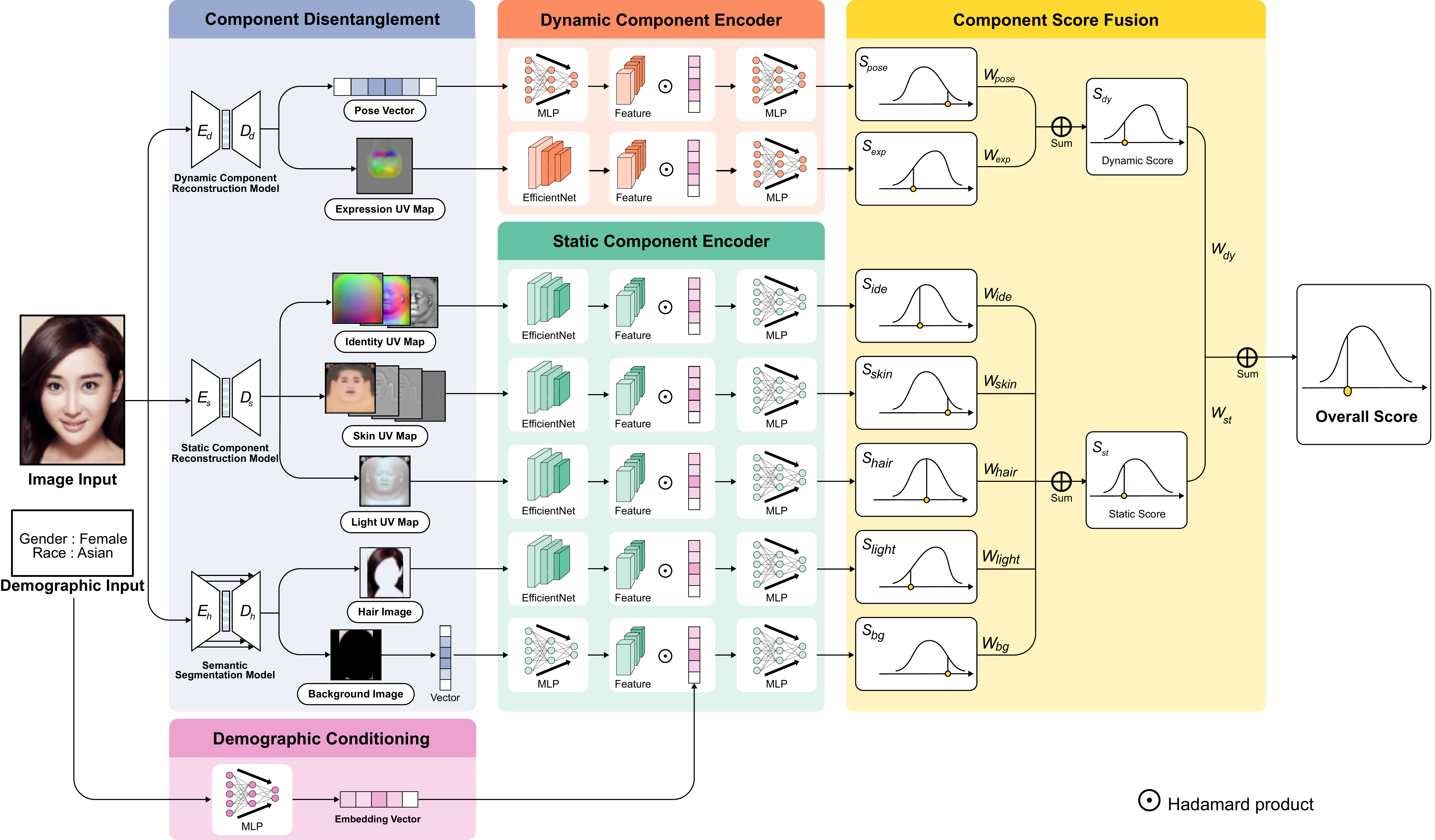}
  \caption{Architecture of DisFace3DNet. Component Disentanglement provides designated inputs to the Static and Dynamic Component Encoders, which produce five nonnegative scores and signed expression and pose scores, respectively. Component Score Fusion yields the overall rating and per-image weighted contributions. Demographic Conditioning provides a shared embedding to the component-specific feature-wise linear modulation (FiLM)~\cite{perez2018film} layers.}
  \label{fig:architecture}
\end{figure*}

DisFace3DNet takes a face image and its benchmark-group label and returns an overall estimate, seven scores, and their weighted contributions (Fig.~\ref{fig:architecture}). Component Disentanglement prepares inputs for the Static and Dynamic Component Encoders; Component Score Fusion combines their outputs. Demographic Conditioning supplies the shared embedding $g$ (Section~\ref{sec:group_conditioning}).

Here, 3D component disentanglement links each designated input to a component-specific branch and its prediction term. For a single image, identity, skin, and hair describe structure and appearance, while light and background describe capture context. These five static scores form a nonnegative base; signed expression and pose scores adjust it for facial state and view.

\subsection{Component Disentanglement}\label{sec:component_disentanglement}
The seven inputs are cached before attractiveness training. The FFHQ-UV pipeline~\cite{bai2023ffhquv}, using Deep3DFaceRecon~\cite{deng2019deep3dface} and the HiFi3D++ topology~\cite{chai2022realy}, provides neutral geometry, albedo, and spherical-harmonic lighting. 3DDFA-V3~\cite{wang2024geometric3ddfa} provides pose and expression, and a 19-class BiSeNet~\cite{yu2018bisenet} provides hair and background masks. Complete input-construction details are provided in Section~\ref*{supp:sec:s1} of the supplemental materials.

\subsubsection{Identity, Skin, and Light}
\emph{Identity} represents expression-neutral facial structure with nine UV channels: three normals, three positions, and three curvatures. Three symmetry and four scale descriptors augment the encoded geometry without supplying skin texture. \emph{Skin} concatenates intrinsic RGB albedo with fine-, middle-, and coarse-scale RGB detail from the unwrapped texture into 12 channels, excluding shape coordinates. \emph{Light} evaluates 27 spherical-harmonic coefficients on neutral-mesh normals to produce three UV channels.

\subsubsection{Pose and Expression}
\emph{Expression} uses expressive and neutral Basel Face Model (BFM) meshes~\cite{paysan2009face} with shared identity coefficients $\boldsymbol\alpha$. Expression coefficients are $\boldsymbol\delta$ for the expressive mesh and zero for the neutral mesh:
\begin{equation}
\Delta V=V_{\mathrm{exp}}(\boldsymbol\alpha,\boldsymbol\delta)-V_{\mathrm{neu}}(\boldsymbol\alpha,\mathbf0).
\end{equation}
Transferring this displacement to HiFi3D++ and UV space yields three displacement-normal channels, three displacement channels standardized and compressed with a hyperbolic tangent, and one log-magnitude channel. \emph{Pose} uses the original-image pitch--yaw--roll vector. Together, these inputs recover state and view information suppressed by neutral frontal fitting.

\subsubsection{Hair and Background}
\emph{Hair} retains segmented RGB hair pixels, with non-hair pixels zeroed before square padding and resizing. To summarize the surrounding scene, \emph{Background} uses a 15-dimensional vector of color, texture, area, and skin-contrast statistics. All routes also receive the supplied group embedding.

\subsection{Benchmark-Group Conditioning} \label{sec:group_conditioning}
Demographic Conditioning in Fig.~\ref{fig:architecture} maps the supplied benchmark-group label to an eight-dimensional embedding $g$ shared by all branches. Before score prediction, each branch applies its own feature-wise linear modulation (FiLM)~\cite{perez2018film} transformation,
\begin{equation}
  \operatorname{FiLM}_k(f_k,g)
  =\bigl(1+\gamma_k(g)\bigr)\odot f_k+\eta_k(g),
\end{equation}
where $f_k$ is component $k$'s feature vector, $\gamma_k$ and $\eta_k$ map $g$ to feature-wise scales and offsets, and $\odot$ denotes element-wise multiplication. This lets each branch condition its score on the benchmark group while sharing its encoder across groups.

\subsection{Static Component Encoder} \label{sec:static_encoder}
Four EfficientNet-B0~\cite{tan2019efficientnet} encoders separately process the Identity, Skin, and Light UV Maps and Hair Image; an MLP encodes the 15-dimensional background vector. Each pooled feature passes through FiLM~\cite{perez2018film} and a score-prediction MLP; the identity route appends symmetry and scale descriptors. Sigmoids produce $S_{\mathrm{ide}}$, $S_{\mathrm{skin}}$, $S_{\mathrm{light}}$, $S_{\mathrm{hair}}$, and $S_{\mathrm{bg}}$.

\subsection{Dynamic Component Encoder} \label{sec:dynamic_encoder}
Expression and pose can have signed associations with attractiveness across facial states and views~\cite{kalayci2014automatic,weng2021twostream,marshall2020human}. An MLP encodes the 3D Pose Vector, and an EfficientNet-B0~\cite{tan2019efficientnet} encodes the Expression UV Map. FiLM~\cite{perez2018film} and a final MLP produce $S_{\mathrm{pose}}$ and $S_{\mathrm{exp}}$.

\subsection{Component Score Fusion} \label{sec:score_fusion}
During representation learning, positive normalized coefficients combine the static scores $S_k(x_k,g)\in[0,1]$. For $\mathcal{K}_{\mathrm{st}}=\{\mathrm{ide},\mathrm{skin},\mathrm{hair},\mathrm{light},\mathrm{bg}\}$, the static score is
\begin{equation}
\begin{aligned}
  S_{\mathrm{st}}
  &=\sum_{k\in\mathcal{K}_{\mathrm{st}}}W_k S_k+b_{\mathrm{st}}, \\
  W_k
  &=\frac{\operatorname{softplus}(\alpha_k)}
  {\sum_{j\in\mathcal{K}_{\mathrm{st}}}\operatorname{softplus}(\alpha_j)} .
\end{aligned}
\end{equation}
Expression and pose use separate learned positive scales on signed readouts,
\begingroup\allowdisplaybreaks[4]
\begin{align}
  S_{\mathrm{exp}}&=W_{\mathrm{exp}}\tanh h_{\mathrm{exp}}(x_{\mathrm{exp}},g),\\
  S_{\mathrm{pose}}&=W_{\mathrm{pose}}\tanh h_{\mathrm{pose}}(x_{\mathrm{pose}},g),\\
  S_{\mathrm{dy}}&=S_{\mathrm{exp}}+S_{\mathrm{pose}}.
\end{align}
\endgroup
The training score $\tilde y_n=S_{\mathrm{st}}+S_{\mathrm{dy}}$ lets overall-rating supervision jointly shape the component branches; $n$ denotes the normalized rating scale.

After freezing the network, we standardize the static and dynamic groups to calibrate their relative influence while retaining the learned static-component proportions. We use their inner-training means $\mu_\ell$ and standard deviations (SDs) $\sigma_\ell$:
\begin{equation}
  Z_{\ell,i}=(S_{\ell,i}-\mu_\ell)/\sigma_\ell,
  \qquad \ell\in\{\mathrm{st},\mathrm{dy}\}.
\end{equation}
The normalized score before clipping is
\begin{equation}
  \hat y^{\mathrm{pre}}_{n,i}=b_0+W_{\mathrm{st}}Z_{\mathrm{st},i}
  +W_{\mathrm{dy}}Z_{\mathrm{dy},i}.
  \label{eq:score_assembly}
\end{equation}
The final fit estimates these three coefficients from the inner-training set $\mathcal{T}$ by minimizing
\begin{equation}
\mathcal{L}_{\mathrm{fit}}=
\sum_{i\in\mathcal{T}}
\left[(\hat y^{\mathrm{pre}}_{n,i}-y_i)^2
+0.3(\hat y^{\mathrm{pre}}_{n,i}-t_{n,i})^2\right],
\label{eq:frozen_fusion_fit}
\end{equation}
subject to $W_{\mathrm{st}},W_{\mathrm{dy}}\geq0$, with $b_0$ unconstrained. Here, $r_i$ is the benchmark mean rating and $t_i$ is the fold-safe FPEM~\cite{li2025facial} prediction for image $i$; $y_i=(r_i-1)/4$ and $t_{n,i}=(t_i-1)/4$ normalize these ratings. This deterministic convex fit updates only fusion coefficients. The frozen network, statistics, and coefficients produce held-out ratings as $\operatorname{clip}(1+4\hat y^{\mathrm{pre}}_n,1,5)$.

Expanding the fitted score gives component terms for $k\in\mathcal{K}_{\mathrm{st}}$ and $d\in\{\mathrm{exp},\mathrm{pose}\}$:
\begin{equation}
 C_{i,k}=\frac{W_{\mathrm{st}}W_kS_{i,k}}{\sigma_{\mathrm{st}}}, \qquad
 C_{i,d}=\frac{W_{\mathrm{dy}}S_{i,d}}{\sigma_{\mathrm{dy}}}.
\end{equation}
The effective intercept collects the fixed offsets:
\begin{equation}
b_{\mathrm{eff}}=b_0+\frac{W_{\mathrm{st}}(b_{\mathrm{st}}-\mu_{\mathrm{st}})}{\sigma_{\mathrm{st}}}-\frac{W_{\mathrm{dy}}\mu_{\mathrm{dy}}}{\sigma_{\mathrm{dy}}}.
\end{equation}
The vector $\mathbf c_i=[C_{i,\mathrm{ide}},\allowbreak C_{i,\mathrm{skin}},\allowbreak C_{i,\mathrm{hair}},\allowbreak C_{i,\mathrm{light}},\allowbreak C_{i,\mathrm{bg}},\allowbreak C_{i,\mathrm{exp}},\allowbreak C_{i,\mathrm{pose}}]$ satisfies $\hat y^{\mathrm{pre}}_{n,i}=b_{\mathrm{eff}}+\sum_k C_{i,k}$. Each $C_{i,k}$ contributes $4C_{i,k}$ rating points before clipping.

\subsection{Joint Learning and Frozen Fusion Fitting} \label{sec:training_scheme}
\begin{figure*}[!t]
\centering
\includegraphics[width=\textwidth]{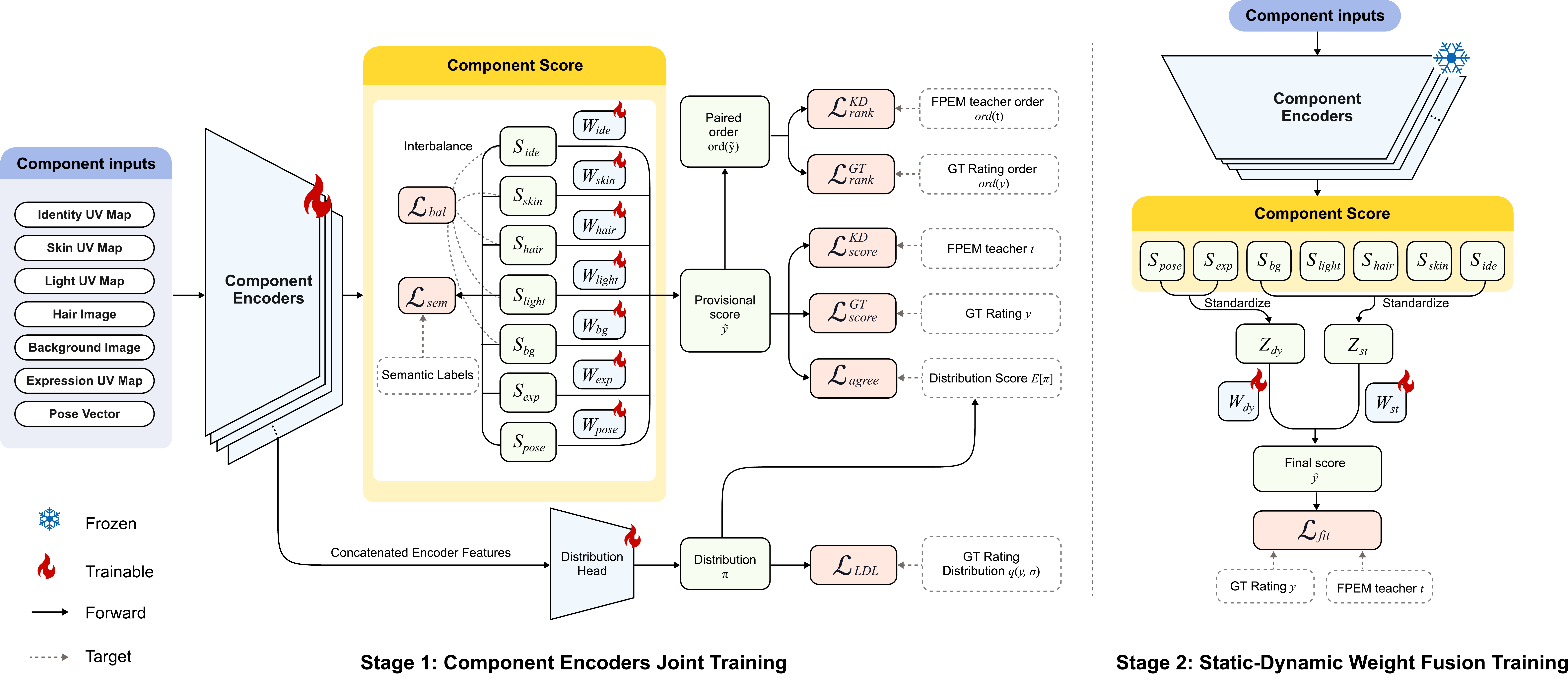}
  \caption{Two-stage training in DisFace3DNet. Stage~1 jointly trains seven component routes using ground-truth (GT) ratings, scalar knowledge distillation (KD) from FPEM~\cite{li2025facial}, and GPT-5.5 weak labels~\cite{openai2026gpt55}. The provisional score $\tilde y_n$ and distribution head $\pi$ receive the indicated loss terms. Stage~2 freezes the network and fits fusion coefficients on the standardized static/dynamic outputs through $\mathcal{L}_{\mathrm{fit}}$.}
\label{fig:joint_training}
\end{figure*}

As shown in Fig.~\ref{fig:joint_training}, Stage~1 (Component Encoders Joint Training) jointly optimizes all component routes, heads, static-component weights, and dynamic scales in shared batches. After inner-validation checkpoint selection, Stage~2 (Static--Dynamic Weight Fusion Training) fits only fusion coefficients on frozen, standardized static/dynamic outputs using inner-training data (Section~\ref{sec:score_fusion}).

Benchmark ratings and scalar FPEM targets supervise the provisional score~\cite{li2025facial,Hinton2015Distilling}; teacher targets receive uniform weights across samples. A 21-bin distribution head predicts $\pi_i$ against a smoothed Gaussian target $q_i$ derived from the benchmark mean and SD. Label-distribution learning (LDL) uses Kullback--Leibler (KL) divergence~\cite{fan2018label,kullback1951information}; agreement aligns the scalar and distribution means. For batch size $B$,
\begingroup\allowdisplaybreaks[4]
\begin{align}
\mathcal L_{\mathrm{score}}^{\mathrm{GT}}&=B^{-1}\sum_i|\tilde y_{n,i}-y_i|,\\
\mathcal L_{\mathrm{score}}^{\mathrm{KD}}&=B^{-1}\sum_i|\tilde y_{n,i}-t_{n,i}|,\\
\mathcal L_{\mathrm{LDL}}&=B^{-1}\sum_{i,m}q_{im}\log(q_{im}/\pi_{im}),\\
\bar y_{\pi,i}&=(\sum_m c_m\pi_{im}-1)/4,\\
\mathcal L_{\mathrm{agree}}&=B^{-1}\sum_i|\tilde y_{n,i}-\operatorname{sg}(\bar y_{\pi,i})|.
\end{align}
\endgroup
Here, $c_m$ is a bin center and $\operatorname{sg}$ stops gradients: LDL trains the distribution head, while agreement trains the component-score path. GT and teacher ordering losses~\cite{lin2019regression,lyu2024ranking} use random batch pairs whose absolute normalized target difference exceeds 0.15; both use a margin of 0.1.

\paragraph{Weak semantic guidance}
Weak semantic labels guide what each branch scores. In our experiments, all 5,500 SCUT-FBP5500 images have seven jointly generated GPT-5.5~\cite{openai2026gpt55} weak labels on a 1--5 ordinal scale from the OpenAI API. We map static labels to $[0,1]$ and center expression/pose labels at neutral on $[-0.10,0.10]$/$[-0.05,0.05]$. These ranges define training targets; standardization and fitted fusion determine their contributions to the final rating. For valid targets $v_{ik}$,
\begin{equation}
  \mathcal{L}_{\mathrm{sem}}
  =\frac{1}{7}\sum_k\frac{1}{N_k}\sum_i M_{ik}|S_{ik}-v_{ik}|.
\end{equation}
Here, $M_{ik}$ selects observed labels and $N_k$ counts valid batch targets for component $k$. These weak labels orient component branches under primary overall-rating supervision. The full prompts and teacher protocol are provided in Section~\ref*{supp:sec:s2} of the supplemental materials.

\paragraph{Static-score balance}
Overall-rating supervision can be satisfied even when a static branch produces nearly constant scores. To encourage variation across all five static branches, we use
\begin{equation}
  \mathcal{L}_{\mathrm{bal}}
  =\operatorname{Var}_{k\in\mathcal{K}_{\mathrm{st}}}
  \left[\log\bigl(\operatorname{Std}_{i}(S_{ik})+0.01\bigr)\right].
\end{equation}
The loss reduces differences in static-score dispersion within a batch. Expression and pose learn their signed scores through semantic, score, KD, agreement, and ordering supervision.

The complete joint objective is
\begin{equation}
\begin{aligned}
\mathcal{L}={}&\mathcal{L}_{\mathrm{score}}^{\mathrm{GT}}
+\lambda_{\mathrm{KD}}\mathcal{L}_{\mathrm{score}}^{\mathrm{KD}}
+\mathcal{L}_{\mathrm{LDL}}
+\lambda_{\mathrm{agree}}\mathcal{L}_{\mathrm{agree}}\\
&+\lambda_{\mathrm{sem}}\mathcal{L}_{\mathrm{sem}}
+\lambda_{\mathrm{bal}}\mathcal{L}_{\mathrm{bal}}\\
&+\lambda_{\mathrm{rank}}(\mathcal{L}_{\mathrm{rank}}^{\mathrm{GT}}
+\mathcal{L}_{\mathrm{rank}}^{\mathrm{KD}}).
\end{aligned}
\label{eq:training_objective}
\end{equation}
FPEM~\cite{li2025facial}, weak labels, and the distribution head are used only for training; group embeddings remain active in deployed FiLM~\cite{perez2018film}.

\section{Experiments} \label{sec:experiments}

\subsection{Dataset}

SCUT-FBP5500~\cite{Liang2018SCUT} contains 5,500 facial images with mean human attractiveness ratings on a 1--5 scale. It provides landmarks, rating distributions, and four benchmark-group labels: Asian male (AM; 2,000 images), Asian female (AF; 2,000 images), Caucasian male (CM; 750 images), and Caucasian female (CF; 750 images). In this evaluation, the learned scores characterize mean human ratings within the population and rating context represented by SCUT-FBP5500~\cite{langlois2000maxims,sutherland2018facial}.

The publisher's fixed five-fold protocol~\cite{Liang2018SCUT} allocates 4,400 training and 1,100 test images per split. DisFace3DNet and matched controls share folds and paired seeds, with a deterministic 3,960/440 inner-training/validation split. Each of the 5,500 images is evaluated once by a model trained on the other folds.

\subsection{Evaluation Metrics}\label{sec:evaluation_metrics}
We evaluate overall predictions against mean human ratings using the Pearson correlation coefficient (PCC), Spearman rank correlation coefficient (SRCC), mean absolute error (MAE), and root mean square error (RMSE). PCC and SRCC measure linear and rank association; MAE and RMSE measure prediction error on the 1--5 rating scale. Higher correlations and lower errors indicate better prediction.

\begin{figure*}[!t]
\begingroup
\makeatletter\def\@captype{table}\makeatother
\centering
\small
\caption{Overall attractiveness prediction on SCUT-FBP5500~\cite{Liang2018SCUT}, grouped by model category}
\label{tab:main_performance}
\setlength{\tabcolsep}{5pt}
\renewcommand{\arraystretch}{1.05}
\begin{tabular*}{\textwidth}{@{\extracolsep{\fill}}lclrrrr@{}}
\toprule
\multirow{2}{*}{\textbf{Model}} & \multirow{2}{*}{\textbf{Year}} & \multirow{2}{*}{\textbf{Backbone}} & \multicolumn{2}{c}{\textbf{Correlation} $\uparrow$} & \multicolumn{2}{c}{\textbf{Error} $\downarrow$} \\
\cmidrule(lr){4-5}\cmidrule(l){6-7}
 & & & \multicolumn{1}{c}{PCC} & \multicolumn{1}{c}{SRCC} & \multicolumn{1}{c}{MAE} & \multicolumn{1}{c}{RMSE} \\
\midrule
\rowcolor{black!6}
\multicolumn{7}{l}{\rule{0pt}{2.1ex}\textbf{CNN-based learning}} \\
Inception-V3~\cite{szegedy2016inception} & 2016 & Inception-V3 & 0.918 & 0.905 & --- & --- \\
ResNeXt-50~\cite{xie2017resnext}& 2017 & ResNeXt-50 & 0.911 & 0.899 & --- & --- \\
DALDL~\cite{chen2019facial} & 2019 & ResNeXt-50 & 0.920 & --- & 0.200 & 0.269 \\
R3CNN~\cite{lin2019regression} & 2019 & ResNeXt-50 & 0.914 & --- & 0.212 & 0.280 \\
NAS4FBP~\cite{zhang2022nas4fbp} & 2022 & Custom & 0.939 & --- & 0.180 & 0.238 \\
Dynamic ER-CNN~\cite{bougourzi2022cnn} & 2023 & ResNeXt-50 + Inception-V3 & 0.921 & --- & 0.204 & 0.267 \\
REX-INCEP~\cite{bougourzi2022dynamic} & 2022 & ResNeXt-50 + Inception-V3 & 0.917 & 0.907 & --- & --- \\
Loss ensemble~\cite{saeed2023automatic} & 2023 & FIAC-Net~\cite{saeed2022fiacnet} & 0.931 & --- & 0.203 & 0.261 \\
RankNet~\cite{lyu2024ranking} & 2024 & ResNet-50 & 0.928 & --- & 0.191 & 0.255 \\
\rowcolor{black!6}
\multicolumn{7}{l}{\rule{0pt}{2.1ex}\textbf{Transformer-based learning}} \\
ViT-B~\cite{dosovitskiy2021vit}& 2021 & ViT-B & 0.880 & 0.869 & --- & --- \\
FBPFormer~\cite{liu2023fbpformer} & 2023 & ViT & 0.918 & --- & 0.205 & 0.273 \\
\rowcolor{black!6}
\multicolumn{7}{l}{\rule{0pt}{2.1ex}\textbf{Attention-based learning}} \\
DyAttenConv~\cite{sun2024dynamic} & 2024 & ResNet-50 & 0.906 & --- & 0.220 & 0.295 \\
\rowcolor{black!6}
\multicolumn{7}{l}{\rule{0pt}{2.1ex}\textbf{Semi-supervised learning}} \\
NFME~\cite{dornaika2020graphbased} & 2020 & VGG-Face & 0.866 & --- & 0.268 & 0.346 \\
FME~\cite{dornaika2022multiview} & 2022 & VGG-Face + ResNet-50 & 0.911 & --- & 0.221 & 0.287 \\
\rowcolor{black!6}
\multicolumn{7}{l}{\rule{0pt}{2.1ex}\textbf{Lightweight learning}} \\
CoAttention~\cite{shi2019improving} & 2019 & MobileNetV2 & 0.926 & 0.916 & 0.202 & 0.266 \\
RIRSCA~\cite{cao2020deep} & 2020 & Custom & 0.878 & --- & 0.252 & 0.332 \\
2D-FAP (60/40)~\cite{li2025facial} & 2025 & MobileNetV2 & 0.915 & 0.903 & --- & --- \\
2D-FAP (five-fold)~\cite{liu2024lightweight} & 2025 & MobileNetV2 & 0.928 & --- & 0.196 & 0.259 \\
\rowcolor{black!6}
\multicolumn{7}{l}{\rule{0pt}{2.1ex}\textbf{Attribute-aware learning}} \\
AaNet~\cite{Lin2019AaNet} & 2019 & ResNet-18 & 0.906 & --- & 0.224 & 0.295 \\
P-AaNet~\cite{Lin2019AaNet} & 2019 & ResNet-18 & 0.897 & --- & 0.229 & 0.304 \\
\rowcolor{black!6}
\multicolumn{7}{l}{\rule{0pt}{2.1ex}\textbf{Multi-task learning}} \\
HMTNet~\cite{xu2019hierarchical} & 2019 & Custom & 0.878 & --- & 0.250 & 0.326 \\
BranchMTL~\cite{vahdati2020facial} & 2020 & ResNet-50 & 0.937 & --- & 0.183 & 0.242 \\
BSN~\cite{li2024broad} & 2024 & ResNeXt-50 & 0.926 & --- & 0.198 & 0.263 \\

\rowcolor{black!6}
\multicolumn{7}{l}{\rule{0pt}{2.1ex}\textbf{Multimodal learning}} \\
FPEM~\cite{li2025facial} & 2025 & Swin-T + CLIP & \textcolor{red}{0.937} & \textcolor{red}{0.932} & \textcolor{red}{0.185} & \textcolor{red}{0.242} \\
\rowcolor{black!6}
\multicolumn{7}{l}{\rule{0pt}{2.1ex}\textbf{3D component-based learning}} \\
\textbf{DisFace3DNet (Ours)} & 2026 & EfficientNet-B0 + MLP & \textbf{0.890} & \textbf{0.877} & \textbf{0.258} & \textbf{0.337} \\
\bottomrule
\end{tabular*}
\endgroup
\end{figure*}

\begin{figure*}[t!]
\centering
\captionsetup{skip=0pt}
\includegraphics[width=\textwidth,trim=0bp 0bp 0bp 4bp,clip]{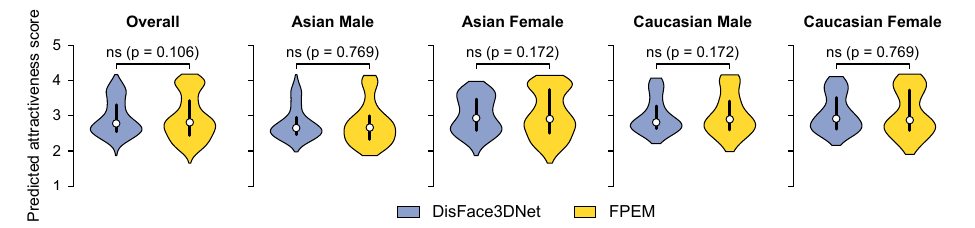}
\caption{DisFace3DNet and FPEM~\cite{li2025facial} prediction distributions for 1,000 held-out images. White dots and black bars indicate medians and interquartile ranges. Paired $t$-test $p$-values are unadjusted overall and Holm-adjusted across the four groups; ns denotes $p\geq0.05$.}
\label{fig:prediction_distributions}
\includegraphics[width=\textwidth]{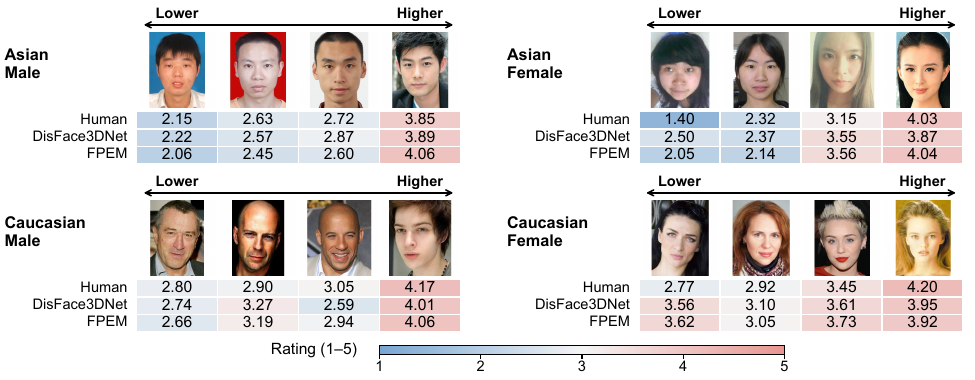}
\caption{Qualitative prediction examples. Four images per group are ordered by increasing human mean rating. Blue/red encodes lower/higher ratings on a fixed 1--5 scale.}
\label{fig:prediction_examples}
\end{figure*}

\subsection{Implementation Details}\label{sec:implementation}
\subsubsection{Input Preparation}
All UV/image inputs use $384\times384$ resolution; pose and background are encoded as vectors. Section~\ref*{supp:sec:s1} of the supplemental materials gives input-construction, normalization, and augmentation details.

\subsubsection{Training Configuration}
EfficientNet-B0~\cite{tan2019efficientnet} encoders use ImageNet initialization~\cite{deng2009imagenet}. We train each fold independently in PyTorch~\cite{paszke2019pytorch} with AdamW~\cite{loshchilov2019adamw}, batch size 12, 70 joint epochs at $3\times10^{-4}$, and 30 refinement epochs at $3\times10^{-5}$.
The semantic coefficient is prespecified as $\lambda_{\mathrm{sem}}=0.2$ using inner-validation prediction and component diagnostics. Inner-validation PCC selects the checkpoint; Stage~2 weight fusion training uses only inner-training data. All choices, standardization statistics, and fitted coefficients are fixed before test-fold evaluation. Layer widths, initialization rules, loss weights, and inference settings are given in Section~\ref*{supp:sec:s1} of the supplemental materials.

\subsection{Baselines}
Inception-V3~\cite{szegedy2016inception} and ResNeXt-50~\cite{xie2017resnext} provide convolutional reference models, while REX-INCEP~\cite{bougourzi2022dynamic} combines their features in a two-branch architecture. ViT-B~\cite{dosovitskiy2021vit} represents transformer-based prediction. 2D-FAP~\cite{liu2024lightweight} uses a lightweight network with dual label distributions. FPEM~\cite{li2025facial} is a state-of-the-art FAP method that combines face-prior and vision--language features, making it a strong reference for overall prediction.

\section{Results}

\subsection{Overall Score Prediction}

FPEM achieves higher correlations and lower errors than DisFace3DNet (Table~\ref{tab:main_performance}). Its face-prior and vision--language features support flexible overall prediction. DisFace3DNet restricts each branch to designated 3D, segmented, or statistical inputs and combines seven scores additively. These constraints trade predictive flexibility for a decomposition that can be inspected component by component.

To compare average prediction levels, Fig.~\ref{fig:prediction_distributions} tests mean differences between DisFace3DNet and FPEM. No significant difference is detected overall or within groups after Holm correction. Figure~\ref{fig:prediction_examples} compares individual predictions with human ratings across the four groups, showing close estimates as well as rating and ordering errors. Section~\ref*{supp:sec:s5} of the supplemental materials gives the sampling protocol and statistical analyses.

\subsection{Component Score Prediction}\label{sec:component_results}

\begin{figure*}[!t]
\centering
\captionsetup{skip=0pt}  % Removes space between figure and caption
\includegraphics[width=\textwidth]{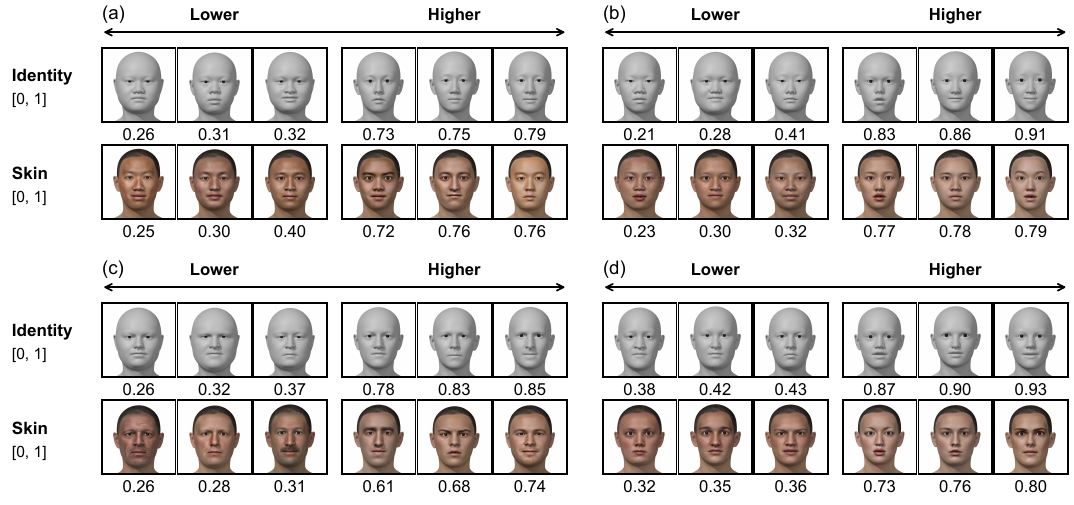}
  \caption{3D identity and skin examples: (a) Asian Male, (b) Asian Female, (c) Caucasian Male, (d) Caucasian Female. Each row shows three lower-score and three higher-score heads. Identity displays neutral geometry; skin uses personal-albedo-guided color and FFHQ-UV detail for display. Labels give raw component scores on $[0,1]$.}
\label{fig:identity_skin_3d}
\end{figure*}

\begin{figure*}[t!]
  \centering
     \captionsetup{skip=0pt}  % Removes space between figure and caption
\includegraphics[width=\textwidth,trim=0bp 19bp 0bp 7bp,clip]{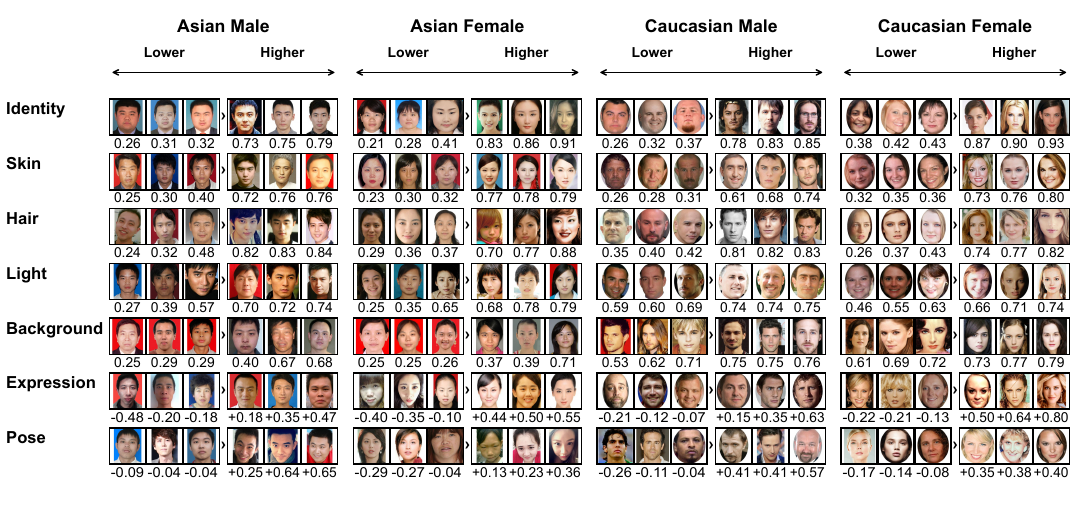}
  \caption{Component-score examples by benchmark group (columns), with three lower- and three higher-score images per component. Labels show $S_k$ for static components ($[0,1]$), $10S_k$ for expression, and $100S_k$ for pose. Displayed expression/pose ranges are $[-0.48,+0.80]$/$[-0.29,+0.65]$.}
  \label{fig:component_output_examples}
\end{figure*}

\subsubsection{Qualitative Component Analysis}
Figure~\ref{fig:identity_skin_3d} shows 48 lower- and higher-score identity and skin heads. Neutral gray identity heads reveal facial contours and feature proportions. Higher-score skin examples generally look more uniform; lower-score examples show stronger local color and texture variation. For visualization, skin heads combine personal-albedo-guided low-frequency color with FFHQ-UV~\cite{bai2023ffhquv} detail under shared rendering conditions.

Figure~\ref{fig:component_output_examples} extends this comparison to 168 examples covering all seven components. Hair rows contrast silhouette and framing, while light and background rows show differences in capture context. Expression and pose rows associate signed scores with eye/mouth appearance and head orientation.

\begin{figure*}[!t]
  \centering
  \captionsetup{skip=0pt}  % Removes space between figure and caption
\includegraphics[width=\textwidth,trim=0bp 11bp 0bp 4bp,clip]{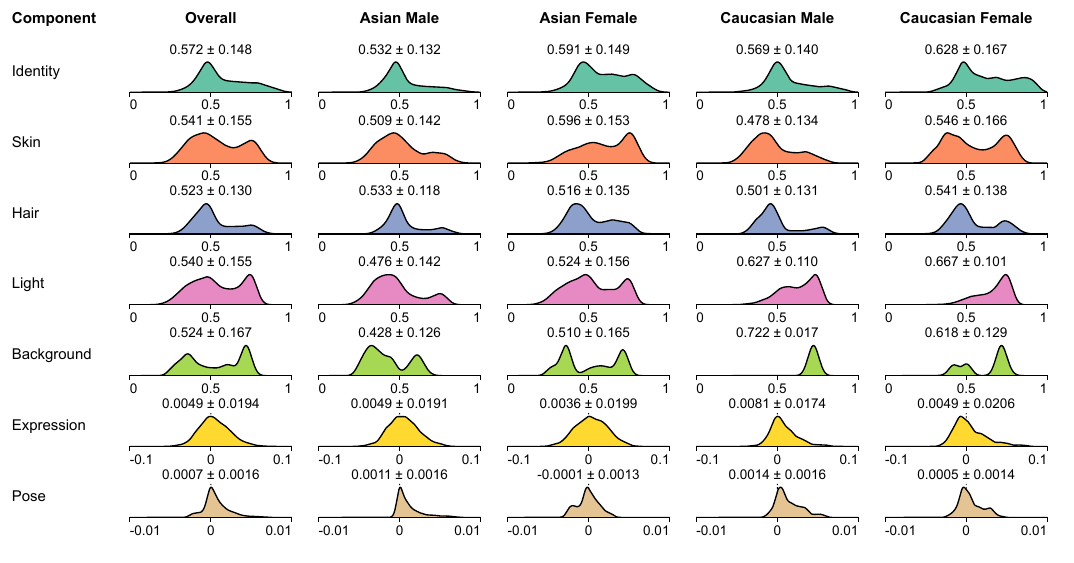}
  \caption{Component-score distributions for all 5,500 held-out SCUT-FBP5500 images and the four benchmark groups. Static scores lie in $[0,1]$; expression and pose use signed axes. Density curves have equal peak heights, and labels report mean $\pm$ sample SD. Scores precede fold-specific standardization and fusion.}
  \label{fig:component_score_distributions}
\end{figure*}

\subsubsection{Component Score Distribution Analysis}
Figure~\ref{fig:component_score_distributions} shows broad static-score subranges, whereas expression and pose concentrate near zero, consistent with their roles as signed adjustments. Light and background show the largest static-score group-mean shifts. Shared axes and bandwidths allow direct comparisons of the raw scores $S_k$ across groups; the overall distribution retains the dataset's group proportions.

\subsubsection{Component Contribution Analysis}
\paragraph{Component contribution variation}
For image $i$ in official fold $f(i)$, we measure component variation using centered rating-point contributions:
\begingroup\predisplaypenalty=0
\begin{equation}
\begin{aligned}
d_{ik} &= 4\left(C_{ik}-\overline C_{f(i),k}\right),\\
a_k &= \frac{1}{N}\sum_i |d_{ik}|, \qquad
p_k = 100\frac{a_k}{\sum_j a_j}.
\end{aligned}
\label{eq:contribution_share}
\end{equation}
\endgroup
$N$ counts evaluated images, and $\overline C_{f(i),k}$ is the fold-specific component mean. The mean absolute deviation $a_k$ measures component $k$'s variation in rating points; $p_k$ expresses its share of the sum across components (100\% in total).

For AM, AF, CM, and CF, centering within each fold--group measures variation among images from the same group. The 95\% intervals resample images within fold--group strata with the fitted models fixed (supplemental Section~\ref*{supp:sec:statistics}).

For all 5,500 held-out images, $b_{\mathrm{eff}}+\sum_k C_{i,k}$ matches the normalized prediction within $1.53\times10^{-7}$ maximum absolute error. All predicted ratings lie within $[1,5]$ before clipping, preserving the additive decomposition in the final output.

\paragraph{Relative contribution magnitudes}
Skin, hair, and identity-related facial shape lead prediction variation in every fold; skin has the largest overall share (Table~\ref{tab:component_contributions}; Eq.~\eqref{eq:contribution_share}).

\begin{table*}[tp]
\centering
\footnotesize
\caption{Relative component contribution magnitudes (\%) with 95\% image-bootstrap intervals conditional on the frozen models. Each column sums to 100\% before rounding. Identity denotes facial shape and structure. AM, AF, CM, and CF denote Asian Male, Asian Female, Caucasian Male, and Caucasian Female, respectively}
\label{tab:component_contributions}
\setlength{\tabcolsep}{5pt}
\begin{tabular}{@{}lccccc@{}}
\toprule
Component & Overall & AM & AF & CM & CF \\
\midrule
Identity & 18.22 [17.98, 18.44] & 18.32 [17.84, 18.79] & 19.86 [19.47, 20.23] & 20.96 [20.15, 21.73] & 21.96 [21.33, 22.56] \\
Skin & 34.29 [33.96, 34.63] & 35.85 [35.24, 36.46] & 35.92 [35.36, 36.51] & 36.21 [35.14, 37.29] & 39.71 [38.81, 40.60] \\
Hair & 20.90 [20.55, 21.25] & 22.25 [21.58, 22.92] & 24.20 [23.60, 24.79] & 25.54 [24.36, 26.67] & 24.14 [23.17, 25.09] \\
Light & 6.82 [6.72, 6.92] & 6.92 [6.73, 7.12] & 7.08 [6.90, 7.25] & 6.12 [5.75, 6.51] & 4.34 [4.05, 4.63] \\
Background & 10.71 [10.56, 10.87] & 6.02 [5.65, 6.38] & 3.98 [3.69, 4.28] & 0.91 [0.84, 0.99] & 0.76 [0.70, 0.84] \\
Expression & 8.45 [8.28, 8.63] & 9.99 [9.63, 10.37] & 8.37 [8.10, 8.65] & 9.51 [8.89, 10.18] & 8.52 [8.06, 8.99] \\
Pose & 0.60 [0.59, 0.62] & 0.65 [0.62, 0.68] & 0.60 [0.58, 0.62] & 0.74 [0.70, 0.79] & 0.57 [0.54, 0.60] \\
\bottomrule
\end{tabular}
\end{table*}

\paragraph{Within-group contributions}
The same three components lead within all four groups (Table~\ref{tab:component_contributions}). Background's smaller within-group shares show that its overall variation includes substantial group-mean differences. Light's share is highest in AF and lowest in CF.

\subsection{Human Evaluation}\label{sec:human_evaluations}
\begin{figure*}[!t]
\centering
\captionsetup{skip=2pt}
\includegraphics[width=\textwidth]{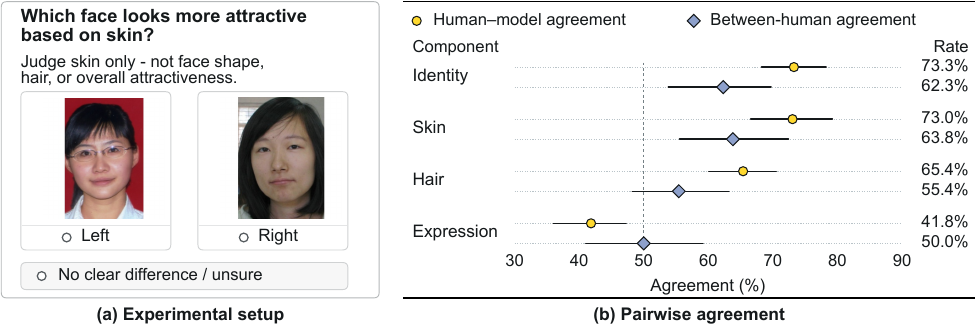}
\caption{Component-specific human evaluation with 20 participants. (a) Restyled skin-trial interface with the original stimulus pair and response options; no response is shown. (b) Pairwise agreement. Yellow circles: fraction of decisive responses matching the model's score ordering. Blue diamonds: mean per-response agreement fraction with other participants on the same component and image pair. Both exclude unsure responses and repeats; diamonds require a decisive peer. Bars: 95\% participant-clustered bootstrap intervals. Vertical line: 50\% agreement between independent fair binary choices.}
\label{fig:human_alignment}
\end{figure*}

Twenty consenting adults (10 male and 10 female) compared randomly sampled image pairs for facial shape (identity), skin, hair, and expression under IRB approval HSEARS20240313013. Responses were Left, Right, or Unsure, with 400 non-repeat responses per component. The post-training study tests whether participants' choices agree with component-score orderings. Figure~\ref{fig:human_alignment}(a) illustrates the interface, and Fig.~\ref{fig:human_alignment}(b) compares human--model and between-human agreement; Section~\ref*{supp:sec:s4} gives the protocol and statistical analysis.

Human--model agreement exceeds the corresponding between-human point estimates for identity, skin, and hair (Fig.~\ref{fig:human_alignment}(b)), supporting these component rankings. Expression shows weak model alignment and low between-human agreement. This task requires comparison with an imagined neutral expression, and the benchmark lacks identity-matched expression variation for controlled validation. Light, background, and pose remain untested in the formal human evaluation; Section~\ref*{supp:sec:s4} explains their omission using informal pilot feedback.

\begin{table*}[!tp]
\centering
\footnotesize
\caption{Five-fold ablations of component scoring, training objectives, and training strategy. Metric summaries are mean $\pm$ sample SD across folds. $\Delta$PCC is DisFace3DNet minus the variant (positive favors DisFace3DNet), with marginal 95\% paired image-bootstrap CIs}
\label{tab:training_signal_ablation}
\setlength{\tabcolsep}{3.0pt}
\begin{tabular}{@{}L{0.22\linewidth}cL{0.25\linewidth}L{0.31\linewidth}@{}}
\toprule
Variant & PCC$\uparrow$ & $\Delta$PCC [95\% CI] & Additional metrics or matched diagnostic \\
\midrule
\multicolumn{4}{@{}l}{\textit{A. Component scoring}} \\
Without component scoring & $0.8931\pm0.0063$ & $-0.0027$ [$-0.0058$, $0.0003$] & SRCC $0.8781\pm0.0066$; MAE $0.2472\pm0.0055$; RMSE $0.3229\pm0.0092$ \\
Without dynamic components & $0.8891\pm0.0043$ & $0.0013$ [$-0.0012$, $0.0038$] & SRCC $0.8752\pm0.0047$; MAE $0.2534\pm0.0050$; RMSE $0.3314\pm0.0062$ \\
Without component-specific inputs & $0.8715\pm0.0079$ & $0.0189$ [$0.0141$, $0.0238$] & Common UV texture for image branches; auxiliary vectors zeroed \\
\midrule
\multicolumn{4}{@{}l}{\textit{B. Training objectives}} \\
Without FPEM~\cite{li2025facial} teacher guidance & $0.8882\pm0.0054$ & $0.0022$ [$-0.0001$, $0.0044$] & SRCC $0.8730\pm0.0052$; MAE $0.2575\pm0.0044$; RMSE $0.3349\pm0.0058$ \\
Without weak semantic supervision ($\lambda_{\mathrm{sem}}=0$) & $0.8896\pm0.0048$ & $0.0008$ [$-0.0015$, $0.0030$] & Lower weak-label correspondence for all seven components in every fold; shifted score distributions \\
\midrule
\multicolumn{4}{@{}l}{\textit{C. Training strategy}} \\
Without joint learning throughout & $0.8875\pm0.0061$ & $0.0028$ [$0.0003$, $0.0053$] & Full staged curriculum; dynamic-score variation and fitted dynamic weight reduced \\
\midrule
\textbf{DisFace3DNet (Ours)} & $0.8904\pm0.0063$ & --- & SRCC $0.8766\pm0.0060$; MAE $0.2580\pm0.0072$; RMSE $0.3369\pm0.0114$ \\
\bottomrule
\end{tabular}
\end{table*}

\begin{figure*}[!t]
\centering
\captionsetup{skip=0pt}  % Removes space between figure and caption
\includegraphics[width=\textwidth]{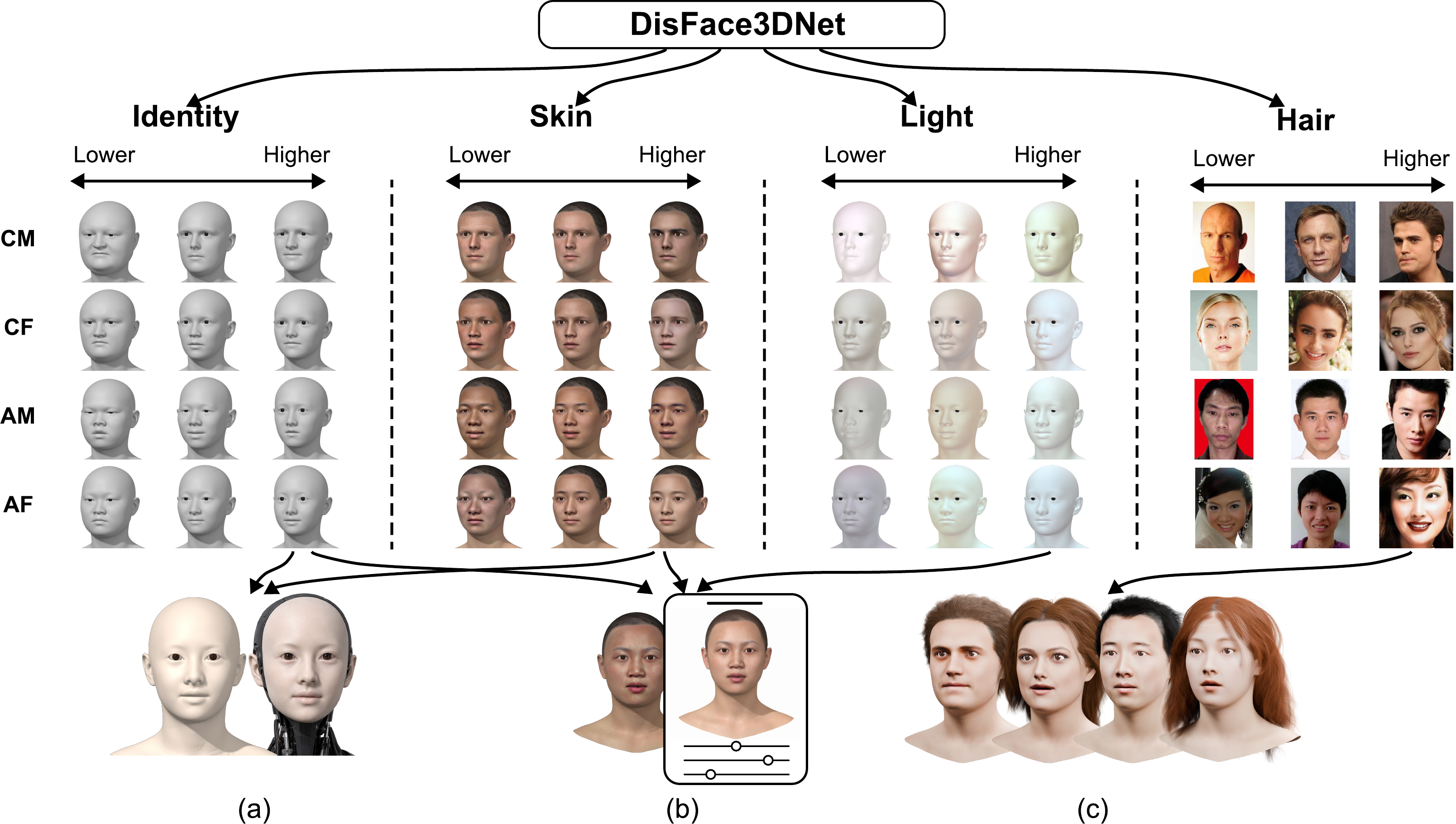}
\caption{Illustrative applications of DisFace3DNet. Component reference scores support (a) digital-human and bio-inspired robot face design, (b) beauty-filter assessment for livestreaming, photography, and social media, and (c) hairstyle design, salon consultation, and wig try-on.}
\label{fig:head_candidate_selection}
\end{figure*}

\subsection{Ablation Studies}
\label{sec:ablations}
Six retrained controls test component scoring, training objectives, and joint learning using overall and component diagnostics (Table~\ref{tab:training_signal_ablation}). Section~\ref*{supp:sec:s3} reports auxiliary controls and teacher diagnostics.

\subsubsection{Effect of Component Scoring}
Replacing component scoring with a capacity-matched scalar head lowers MAE; the paired 95\% CI for its PCC difference includes zero. Thus, the additive scores incur a modest absolute-error cost. Removing dynamic components also gives a PCC interval spanning zero. Replacing component-specific inputs with a common UV texture lowers PCC, supporting the value of organizing inputs by component (Table~\ref{tab:training_signal_ablation}).

\subsubsection{Effect of the Training Objectives}
Removing weak semantic supervision lowers weak-label correspondence for all seven components in every fold despite little change in overall PCC, showing why component correspondence must be assessed alongside overall accuracy. FPEM guidance improves correlations in four folds, but its paired PCC interval includes zero and neither error metric improves.

Five seed-specific sweeps over five semantic coefficients show a trade-off between overall prediction and component correspondence with weak labels. The prespecified $\lambda_{\mathrm{sem}}=0.2$ is retained; Fig.~\ref*{supp:fig:component_supervision} of the supplemental materials gives all distributions and paired seed trajectories.

\subsubsection{Effect of the Training Strategy}
Joint learning outperforms the compute-matched staged curriculum in PCC and retains more dynamic-score variation.

\section{Applications}\label{sec:applications}
Figure~\ref{fig:head_candidate_selection} uses component reference scores to illustrate geometry selection for digital-human and bio-inspired robot faces, beauty-filter assessment, and hairstyle comparison for salon consultation and wig try-on. Section~\ref*{supp:sec:s6} of the supplemental materials details these workflows.

\section{Conclusion}\label{sec:conclusion}
DisFace3DNet predicts overall attractiveness through seven reference scores whose weighted contributions sum to each estimate. On SCUT-FBP5500, skin, hair, and identity lead prediction variation, and human comparisons support their score directions. Expression alignment is weaker, while light, background, and pose have not undergone formal human evaluation. The illustrated workflows show how the scores can organize design references and appearance comparisons.
\par

\AtEndDocument{\newwrite\maincounts\immediate\openout\maincounts=main_counts.tex
\immediate\write\maincounts{\string\setcounter{figure}{\arabic{figure}}}
\immediate\write\maincounts{\string\setcounter{table}{\arabic{table}}}
\immediate\write\maincounts{\string\setcounter{equation}{\arabic{equation}}}
\immediate\closeout\maincounts}
\bibliographystyle{IEEEtran}
\bibliography{mybibfile_ieee,disface_bibstyle}
\end{document}

\bstctlcite{disfaceIEEEcontrol}
\raggedbottom
\title{Supplemental Materials for\\\DisFacePaperTitle}
\author{Fenggui Rao~\orcidlink{0009-0000-8990-1515},
Yan Luximon~\orcidlink{0000-0003-2843-847X},
and Jie Zhang~\orcidlink{0000-0001-8219-5590}
}

\markboth{DisFace3DNet -- Supplemental Materials}{Rao \MakeLowercase{\textit{et al.}}: DisFace3DNet}
\maketitle
\thispagestyle{empty}
\section{Overview}\label{sec:overview}
These materials detail implementation, ablation and sensitivity analyses, human evaluation, prediction comparisons, and application workflows. Figure, table, and equation numbering continues from the main paper.

\section{Implementation and Supervision}\label{sec:method_supplement}
\subsection{Input Construction and Implementation}\label{sec:s1}
The curvature channels in Section~\ref*{main:sec:component_disentanglement} of the main paper contain mean and Gaussian curvature and shape index. Three symmetry descriptors compare UV halves; four standardized scale descriptors are $[\operatorname{Std}(X),\operatorname{Std}(Y),\operatorname{Std}(Z),\operatorname{Range}(Z)]$.

Expression displacement $\Delta V$ is embedded, transferred, and scattered onto HiFi3D++ before UV unwrapping:
\begin{align}
\Delta U&=\mathcal U(\mathcal T_{\mathrm{BFM}\rightarrow\mathrm{HiFi}}(\Delta V)),\\
X_{\mathrm{exp}}&=[\mathcal N(\Delta U),\,D_{\mathrm{exp}},\,M_{\mathrm{exp}}],\\
D_{\mathrm{exp}}&=3\tanh(z(\Delta U)/3),\notag\\
M_{\mathrm{exp}}&=\log(1+100\|\Delta U\|_2).\notag
\end{align}
Here, $\mathcal T$ transfers BFM displacement to HiFi3D++, $\mathcal U$ unwraps it into UV coordinates, and $\mathcal N$ computes three normal channels from raw displacement. We standardize each displacement channel using fixed cached statistics from 500 samples, shared across folds, then compress large values with the hyperbolic tangent and zero invalid pixels. 3DDFA-V3 uses RetinaFace--ResNet50~\cite{deng2020retinaface,he2016resnet} and outputs pitch, yaw, and roll in radians. BiSeNet classes 17 and 0 define hair and background. The background vector summarizes RGB means/SDs (six), gradient-magnitude mean/SD (two), area ratio (one), RGB distance from the skin mean (one), HSV means (three), and saturation/value SDs (two).

\paragraph{Input normalization and augmentation}
Identity positions are standardized over valid pixels, with invalid pixels zeroed. Finite-difference normals from unstandardized positions map from $[-1,1]$ to $[0,1]$. Mean and Gaussian curvature are divided by 5 and 10; shape index keeps its prepared $[-1,1]$ scale. During the first 70 epochs, skin-albedo brightness and contrast factors are independently uniform on $[0.85,1.15]$ with probability 0.3. With probability 0.15, one of hair, light, or all skin-detail channels is uniformly selected and zeroed; identity-normal Gaussian noise has SD 0.02 and probability 0.15. Refinement disables augmentation.

\paragraph{Training configuration}
Each CNN's 1,280 pooled features are projected to 256 dimensions. Background and pose MLP widths are $15$--$64$--$32$ and $3$--$32$--$16$; the expression readout is $256$--$64$--$1$. Identity appends its seven descriptors before scoring. Dropout and stochastic-depth probabilities are 0.3. EfficientNet feature blocks 0--5 remain frozen for RGB inputs and 1--5 for adapted inputs. Adapted stems initialize their first three channels at 0.5 times the pretrained weights and cycle those weights at factor 0.1 for extra channels. FiLM scale/offset mappings start at zero.

Training uses an NVIDIA H200 GPU, weight decay 0.01, mixed precision, and gradient-norm clipping at 5.

For LDL, we normalize a discrete Gaussian over 21 equally spaced bin centers on $[1,5]$, using the rating SD clipped to $[0.20,1.0]$, then mix it with 2\% uniform mass.

In Eq.~(\ref*{main:eq:training_objective}), $\lambda_{\mathrm{KD}}=\lambda_{\mathrm{agree}}=\lambda_{\mathrm{bal}}=0.3$, $\lambda_{\mathrm{sem}}=0.2$, and $\lambda_{\mathrm{rank}}=0.1$; GT-score and LDL weights are one. Model selection and frozen fusion fitting are described in Sections~\ref*{main:sec:implementation} and~\ref*{main:sec:training_scheme} of the main paper.

\paragraph{Inference configuration}
Inference averages overall and component predictions from original and horizontally flipped inputs. Mirroring reverses horizontal geometry and displacement components and yaw. Demographic Conditioning through FiLM~\cite{perez2018film} remains active in both views.

\subsection{Weak Semantic Supervision and Teacher Protocol}\label{sec:s2}
In the reported experiments, the following prompts provided auxiliary semantic guidance by jointly generating seven GPT-5.5 integer labels (1--5) per image for all 5,500 images. Shape denotes the identity component (facial shape and structure). For an integer label $v$, static, expression, and pose targets map to $(v-1)/4$, $(v-3)/20$, and $(v-3)/40$, respectively; Section~\ref*{main:sec:training_scheme} of the main paper defines the masked loss.

\begingroup
\linespread{.98}\selectfont
\clubpenalty=10000
\paragraph{System prompt}
You are a careful image-factor annotator for a released academic facial-image benchmark.

Your task is NOT to identify the person, infer demographics, or make a personal judgment. Treat the image only as a benchmark sample with an existing aggregate face-score. Your task is to estimate how specific visible factors would push that benchmark face-score up or down in the current image.

Definition: The benchmark face-score is the aggregate visual preference / appeal rating that dataset raters would assign to the face image, based only on visible image evidence.

\noindent Important rules:\par
\noindent 1. Evaluate only the requested factor.\\
2. Treat all other factors as fixed as much as possible.\\
3. Judge contribution to the benchmark face-score, not objective quality alone.\\
4. A score of 3 means neutral / ordinary / little clear effect, not bad.\\
5. Do not reward intensity by itself.\\
6. Subtle, harmonious, natural effects may score higher than stronger but less flattering ones.\\
7. Do not let identity, age, gender, ethnicity, clothing, or personal preference dominate the judgment.\\
8. Use 1 or 5 only when the effect is clearly strong.

\paragraph{User prompt}
Evaluate the contribution of the following 7 factors to the benchmark face-score of the face in this image. Rate each factor on a 1-5 integer scale.

\noindent Unified scoring rule for every factor:\par
\noindent 1 = strongly lowers the benchmark face-score\\
2 = slightly lowers the benchmark face-score\\
3 = little or no clear effect / neutral / ordinary\\
4 = slightly raises the benchmark face-score\\
5 = strongly raises the benchmark face-score

Factor definitions:
\nopagebreak[4]

\noindent\textbf{1. Facial Structure Contribution}\par\nopagebreak[4]
\noindent Judge how the face geometry itself contributes to the benchmark face-score: facial proportion, symmetry, contour, spacing, and structural harmony of features. Ignore skin texture, hairstyle, expression, lighting, and background.

\noindent\textbf{2. Skin Contribution}\par\nopagebreak[4]
\noindent Judge how the visible skin contributes to the benchmark face-score: clarity, smoothness, evenness, cleanliness, healthy-looking texture/tone. Ignore facial shape, hairstyle, expression, lighting direction, and background.

\noindent\textbf{3. Hair Contribution}\par\nopagebreak[4]
\noindent Judge how the hairstyle, tidiness, framing, volume, and harmony of the hair with the face contribute to the benchmark face-score. Ignore facial shape, skin quality, expression, lighting, and background.

\noindent\textbf{4. Expression Contribution}\par\nopagebreak[4]
\noindent Compared with a relaxed neutral expression of the same person under the same pose, lighting, and background, judge how much the current facial expression lowers or raises the benchmark face-score. Do not reward smile size or smile intensity by itself. A subtle natural smile may score higher than a wide or exaggerated smile. Judge only the contribution of the expression to attractiveness.

\noindent\textbf{5. Pose Contribution}\par\nopagebreak[4]
\noindent Judge how the head angle, tilt, and direction of gaze contribute to the benchmark face-score. Consider whether the angle is flattering, whether the gaze engages the viewer, and whether the head position looks natural and well-composed. A standard frontal pose with no special flattering effect is 3. A slight natural three-quarter angle or subtle tilt that flatters the face can be 4. An extreme angle, looking away, or very awkward tilt is 1-2. Ignore facial structure, skin quality, hairstyle, expression, lighting, and background.

\noindent\textbf{6. Lighting Contribution}\par\nopagebreak[4]
\noindent Judge how the current lighting pushes the benchmark face-score up or down: shadow placement, softness, balance, visibility of features, and whether the lighting flatters the face. Ignore the person's intrinsic facial structure, skin condition, hairstyle, expression, and background content.

\noindent\textbf{7. Background Contribution}\par\nopagebreak[4]
\noindent Judge how the current background helps or hurts the benchmark face-score: distraction, clutter, color harmony, and whether it supports the face as the visual focus. Ignore the face itself, skin, hair, expression, and lighting quality on the face.

\noindent Calibration examples:\par
\noindent - Example (expression): a slight relaxed smile that flatters the face \texttt{->} 4\\
- Example (expression): a forced or awkward smile that hurts the look \texttt{->} 2\\
- Example (lighting): harsh shadows under the eyes that make the face less flattering \texttt{->} 2\\
- Example (skin): clear, even, healthy-looking skin that supports attractiveness \texttt{->} 4\\
- Example (pose): slight three-quarter turn that shows the jawline well \texttt{->} 4

\noindent Important:\par
\noindent - Judge contribution to the benchmark face-score, not general image quality.\\
- Do not rate the person; annotate visible image factors for research weak supervision.\\
- Do not let the overall face-score dominate your answer.\\
- Keep the seven factors as independent as possible.

First, give one short observation phrase for each factor. Then output your final scores on the last line in EXACTLY this format: SCORES: [shape],\allowbreak[skin],\allowbreak[hair],\allowbreak[expression],\allowbreak[pose],\allowbreak[light],\allowbreak[background]

\endgroup

\paragraph{Teacher}
The FPEM~\cite{li2025facial} teacher follows the student's official folds. Starting from the public checkpoint, it is fine-tuned separately on each training fold, with checkpoint selection by internal validation. We average its original/flip five-class probabilities to obtain the expected scalar rating used for uniformly weighted score distillation, ordering, and frozen fusion fitting. Benchmark rating statistics define the 21-bin LDL target.

\section{Extended Ablations and Sensitivity}\label{sec:s3}

\begin{figure*}[!tp]
\centering
\captionsetup{skip=0pt}  % Removes space between figure and caption
\includegraphics[width=\textwidth]{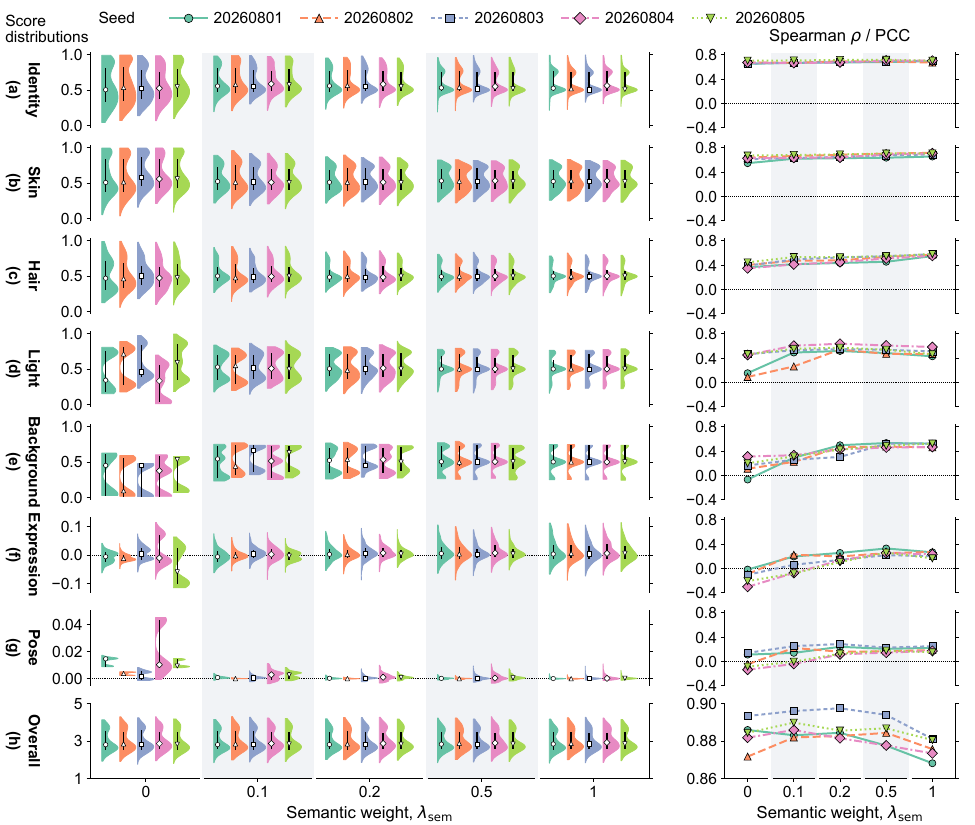}
\caption{Sensitivity to $\lambda_{\mathrm{sem}}$. Half violins (left) show (a--g) component scores and (h) predicted ratings; white-centered markers and black bars mark medians and interquartile ranges. Curves (right) show (a--g) weak-label Spearman correlations and (h) PCC with mean human ratings. Colors identify seeds; curves connect same-seed results.}
\label{fig:component_supervision}
\end{figure*}

\subsection{Statistical Procedures}\label{sec:statistics}
Five-fold metrics and component diagnostics are reported as equally weighted fold means with sample SDs (denominator four). Weak-supervision diagnostics compare raw scores and within-fold weak-label Spearman correlations on the same held-out images. Ablation differences are DisFace3DNet minus the variant. Their 95\% percentile confidence intervals (CIs) use 10,000 paired within-fold image-bootstrap resamples~\cite{efron1979bootstrap}, preserving fold sizes and model pairing. Each resample recomputes fold PCC/MAE and differences between five-fold means.

For Table~\ref*{main:tab:component_contributions}, 10,000 fold--group-stratified image-bootstrap resamples (seed 20260905) recompute component means and shares while preserving group composition. Both procedures keep the trained models fixed and measure uncertainty from image sampling. Within each fold, static-score scale coefficient of variation (CV) is the population SD of the five held-out component SDs divided by their mean.

\subsection{Control Definitions}
All five-fold controls retain the common settings except for the named change.

\paragraph{Component scoring}
Without component scoring, one scalar head replaces the seven component readouts and their additive fusion, with 224 fewer parameters ($<0.001\%$). Inputs, encoders, conditioning, and schedule remain the same. Without dynamic components, the five static routes are retrained. Without component-specific inputs, all image branches receive one UV texture and auxiliary vectors (pose, background, symmetry, and scale) are zeroed; heads, conditioning, and joint training remain.

\paragraph{Training objectives}
Removing teacher guidance excludes FPEM~\cite{li2025facial} scalar and ordering targets from learning and its scalar target from frozen fusion fitting. Removing weak semantic supervision sets $\lambda_{\mathrm{sem}}=0$. Auxiliary controls remove balance loss, distribution consistency (LDL and scalar--distribution agreement), both ordering losses, or the group embedding separately.

\paragraph{Training strategy}
Joint learning optimizes all seven routes throughout both learning-rate periods. The compute-matched staged control learns static routes, then jointly trains expression and pose, and finally refines all routes. Both end with the same frozen fusion fitting.

\subsection{Prediction Error and Component-Score Diagnostics}
\paragraph{Component scoring}
Predicting through seven component scores increases mean MAE by 0.0108 rating points (95\% CI: $0.0078$ to $0.0138$) relative to the capacity-matched scalar control, quantifying the additional error associated with the additive scoring structure.

\paragraph{Weak semantic supervision}
Without weak supervision, expression weak-label correlation falls from $0.116\pm0.067$ to $-0.188\pm0.104$, and pose correlation from $0.212\pm0.035$ to $-0.022\pm0.060$. Expression-score SD changes from $0.0190\pm0.0016$ to $0.0295\pm0.0301$, indicating greater fold-to-fold variation in scale; mean pose score shifts from $0.0007\pm0.0005$ to $0.0082\pm0.0051$. 

\paragraph{Training strategy}
Compared with joint learning, staged training reduces expression-score SD from $0.0190\pm0.0016$ to $0.0023\pm0.0008$ and the fitted dynamic coefficient from $0.0242\pm0.0050$ to $0.0034\pm0.0014$ across held-out folds. The staged model therefore uses a narrower range of expression scores and gives dynamic variation less influence on the final rating.

\paragraph{Teacher guidance}
Adding FPEM guidance changes mean MAE by 0.0006 points (95\% CI $-0.0017$--$0.0028$) and raises within-fold weak-label Spearman correlation from $0.329\pm0.198$ to $0.450\pm0.104$ for background, while lowering it from $0.233\pm0.082$ to $0.116\pm0.067$ for expression and from $0.264\pm0.046$ to $0.212\pm0.035$ for pose.

\subsection{Semantic-Weight Sensitivity}
\subsubsection{Evaluation Design}
Five semantic coefficients are evaluated with five seeds (20260801--20260805), giving 25 model fits on fold~0 inner validation. Within each seed, coefficients share a 3,960/440 partition of the 4,400 official training images. Seeds vary the partition and training randomness; the 1,100-image test fold is unused. In Fig.~\ref{fig:component_supervision}, Scott-bandwidth half violins have equal maximum widths; axes are fixed within components without per-run standardization.

\subsubsection{Results}
As $\lambda_{\mathrm{sem}}$ increases from zero to 0.2, weak-label Spearman correlation improves for all seven components in all five seeds. Averaged over seeds and components, static correlation rises from 0.430 to 0.564, and expression/pose correlation from $-0.077$ to 0.175. All five static-score SDs and interquartile ranges decrease in every seed.

From 0.1 to 0.2, mean weak-label Spearman correlation rises for every component, while PCC falls from $0.8873\pm0.0057$ to $0.8863\pm0.0064$ (Fig.~\ref{fig:component_supervision}(h); mean $\pm$ sample SD across five seeds). At both 0.5 and 1.0, skin, hair, and expression correlations exceed their 0.2 values in every seed, whereas mean light correlation falls from 0.564 at 0.2 to 0.492 at 1.0. Mean PCC decreases to $0.8840\pm0.0069$ at 0.5 and $0.8758\pm0.0053$ at 1.0.

\subsection{Auxiliary-Control Results}
\begin{table*}[!t]
\centering
\footnotesize
\caption{Auxiliary-design ablations. Intervals follow Section~\ref{sec:statistics}. Paired diagnostic values list the variant before DisFace3DNet}
\label{tab:auxiliary_controls}
\setlength{\tabcolsep}{3pt}
\begin{tabular}{@{}L{0.22\linewidth}cL{0.25\linewidth}L{0.31\linewidth}@{}}
\toprule
Variant & PCC$\uparrow$ & $\Delta$PCC [95\% CI] & Matched diagnostic \\
\midrule
Without balance loss & $0.8909\pm0.0069$ & $-0.0005$ [$-0.0028$, $0.0017$] & Static-score scale CV: $0.377\pm0.029$ vs. $0.082\pm0.011$ \\
Without distribution consistency & $0.8893\pm0.0056$ & $0.0011$ [$-0.0016$, $0.0037$] & KL divergence: $0.508\pm0.002$ vs. $0.100\pm0.005$; score--distribution MAE: $0.464\pm0.015$ vs. $0.046\pm0.002$ points \\
Without pairwise ordering & $0.8899\pm0.0055$ & $0.0005$ [$-0.0015$, $0.0024$] & GT-pair concordance: $(96.79\pm0.35)\%$ vs. $(96.81\pm0.32)\%$ \\
Without group conditioning & $0.8885\pm0.0042$ & $0.0019$ [$-0.0003$, $0.0040$] & Group embedding removed \\
\midrule
\textbf{DisFace3DNet (Ours)} & $0.8904\pm0.0063$ & --- & Reference for all comparisons \\
\bottomrule
\end{tabular}
\end{table*}
Table~\ref{tab:auxiliary_controls} shows that balance loss reduces scale CV, making static-branch variability more comparable. Removing distribution consistency worsens both held-out distribution diagnostics in every fold. Without pairwise ordering, SRCC changes from $0.8766\pm0.0060$ to $0.8756\pm0.0067$; the unrounded GT-concordance difference is below 0.02 percentage points. The paired PCC intervals for all four auxiliary controls include zero.

Changing only the supplied AM/AF/CM/CF token in the fixed model produces a mean rating range of 0.074 points (95\% CI 0.060--0.086).

\section{Human Evaluation}\label{sec:s4}
\subsection{Evaluation Design}
Twenty consenting adults (10 male, 10 female) completed 2,060 responses under The Hong Kong Polytechnic University IRB approval HSEARS20240313013. Each completed 103 trials: four practice, 80 non-repeat component, five component repeats, ten separate overall-attractiveness, two identical-image attention, and two instruction checks.

Images were sampled uniformly from SCUT-FBP5500 without group, fold, or target-gap restrictions; left--right order was randomized. Participants compared facial shape (identity), skin, hair, or expression, with an unsure option. The post-training comparisons evaluate whether component-score orderings agree with participants' choices. The Chinese/English interface requested a laptop/desktop display and loaded both images before allowing a response.

Informal pilot feedback guided component selection. Participants found it difficult to isolate lighting's contribution, discern pose differences in the candidate pairs, or apply a clear attractiveness criterion to background differences. The formal study therefore focused on identity, skin, hair, and expression.

Instructions restricted attention to the named attribute, excluded overall attractiveness, and offered ``No clear difference / unsure''. Shape trials excluded skin, hair, and makeup; skin trials excluded shape and hair; hair trials excluded the face. Expression trials asked: ``Whose expression appears to increase that face's attractiveness more compared with a neutral expression?'' Identical-image checks instructed unsure, and instruction checks requested Left. Figure~\ref*{main:fig:human_alignment}(a) in the main paper illustrates a skin trial.

\subsection{Statistical Analysis}
Each component contributes 400 non-repeat responses; the primary agreement estimates exclude unsure responses. For this analysis, held-out component scores are standardized by their official-fold means and sample SDs; model direction follows the higher standardized score. No evaluated pair has tied model scores. Human--model agreement is the fraction of decisive responses following this direction. For between-human agreement, each decisive response is compared with all other participants' decisive responses to the same component and image pair, aligning the chosen image rather than its screen position. We average these per-response agreement fractions, weighting responses equally in both estimates. A response without another decisive response to that pair is omitted only from the between-human estimate.

Both estimates use 95\% percentile intervals from 100,000 participant-clustered bootstrap draws~\cite{efron1979bootstrap} (seed 20260724). Each draw samples 20 participants with replacement, retains their response clusters, and recomputes both estimates. To compare different people, between-human agreement uses responses from other original participants, even when a draw samples the same participant more than once. A response without such a peer on the same pair is omitted from that draw's between-human estimate. For reference, response-level Wilson intervals~\cite{wilson1927probable} for human--model agreement are 68.2\%--77.7\%, 67.7\%--77.8\%, 60.2\%--70.3\%, and 36.3\%--47.5\% for identity, skin, hair, and expression, respectively.

Secondary analyses of identity, skin, and hair examine pairs whose component-score and benchmark-rating directions conflict (excluding equal ratings), count unsure as non-agreement, and compare human majority choices with model directions. Majority analysis excludes unsure and tied votes; each component-specific pair counts once, including unanimous choices.

\subsection{Agreement Counts and Secondary Analyses}
\paragraph{Response counts}
All participants passed the attention and instruction checks. Figure~\ref*{main:fig:human_alignment}(b) reports the four components separately. Human--model agreeing/decisive counts for identity, skin, hair, and expression are 241/329, 214/293, 221/338, and 123/294. Between-human estimates use 329, 291, 338, and 292 responses across 80, 78, 80, and 78 image pairs, respectively: two skin and two expression responses lack another decisive response. There are 1,066, 852, 1,144, and 856 response-to-response comparisons, counting both reference directions; the reported rates average per-response fractions rather than divide aggregate comparison counts.

\paragraph{Pooled static-component agreement}
Pooling responses for identity, skin, and hair gives 676/960 (70.4\%; participant-clustered 95\% CI: 67.1\%--73.7\%).

\paragraph{Conflicting directions, unsure responses, and majority votes}
For the same three components, participants follow the component score in 205/334 decisive conflicting-direction responses (61.4\%). Counting the 240 unsure responses for identity, skin, and hair as non-agreement gives 56.3\% (676/1,200). Human majority choices match the model for 178/214 non-tied pairs from these three components (83.2\%).

\section{Prediction Comparisons with FPEM}\label{sec:s5}

\subsection{Mean-Prediction Comparison Protocol}
For Fig.~\ref*{main:fig:prediction_distributions}, 1,000 held-out images were sampled uniformly without replacement (AM/AF/CM/CF: 370/351/138/141). Two-sided paired $t$-tests compare DisFace3DNet and FPEM predictions; subgroup tests use Holm correction at $\alpha=0.05$.

\subsection{Overall and Group Mean Differences}
On the sampled images, mean predictions are 2.927 (DisFace3DNet) and 2.940 (FPEM); the overall paired test gives $t(999)=-1.619$ and $p=0.106$. The AM/AF/CM/CF mean offsets (DisFace3DNet minus FPEM) are $+0.011/-0.027/-0.040/-0.012$ points.

\subsection{Image-Level Prediction Errors}

The 16 held-out examples in Fig.~\ref*{main:fig:prediction_examples} of the main paper show both shared and model-specific prediction errors. Both models overestimate the first CF and third AF images and reverse the second/third CM ordering. Only DisFace3DNet reverses the first two AF images; FPEM preserves their order but overestimates the first. Several DisFace3DNet estimates closely match human ratings: using unrounded values, absolute errors are 0.06 (second AF), 0.06 (second AM), 0.15 (fourth CM), and 0.04 (fourth AM) points.

\section{Applications}\label{sec:s6}
\paragraph{Digital-Human and Robot Face Design}
A designer could select geometry references by identity score, then compare skin and hair with geometry fixed (Fig.~\ref*{main:fig:head_candidate_selection}(a)). This separates shape and appearance choices.

\paragraph{Beauty-Filter Assessment}
For presets on the same image, users could compare skin, light, and identity scores alongside the overall rating when selecting surface treatments; Fig.~\ref*{main:fig:head_candidate_selection}(b) illustrates this comparison.

\paragraph{Hairstyle Design and Virtual Try-On}
Hair scores could shortlist hairstyle references for transfer to the same face (Fig.~\ref*{main:fig:head_candidate_selection}(c)). Try-on images and user preferences would then guide salon or wig selection.

\FloatBarrier
\bibliographystyle{IEEEtran}
\bibliography{mybibfile_ieee,disface_bibstyle}

%% file: disface_preamble.tex
\usepackage[letterpaper,margin=1in,bottom=76pt,headheight=8pt,headsep=10pt,footskip=0pt]{geometry}
\usepackage{amsmath,amsfonts}
\usepackage{algorithmic}
\usepackage{array}
\usepackage[caption=false,font=normalsize,labelfont=sf,textfont=sf]{subfig}
\usepackage{textcomp}
\usepackage{stfloats}
\usepackage{url}
\usepackage{verbatim}
\usepackage{graphicx}
\usepackage{booktabs}
\usepackage{colortbl,multirow}
\usepackage{placeins}
\def\BibTeX{{\rm B\kern-.05em{\sc i\kern-.025em b}\kern-.08em
    T\kern-.1667em\lower.7ex\hbox{E}\kern-.125emX}}
\newcolumntype{L}[1]{>{\raggedright\arraybackslash}p{#1}}

\usepackage{tikz}
\usepackage{xcolor}

\usepackage{orcidlink}
\usepackage{xr}
\hypersetup{hidelinks}

\IEEEaftertitletext{\vspace{-24pt}}

\newcommand{\DisFacePaperTitle}{DisFace3DNet: Explainable Facial Attractiveness Prediction via 3D Component Disentanglement}

\makeatletter
\def\section{\@startsection{section}{1}{\z@}{1.5ex plus 1.5ex minus 0ex}%
{0.7ex plus 1ex minus 0ex}{\normalfont\normalsize\centering\scshape}}
\def\subsection{\@startsection{subsection}{2}{\z@}{2.0ex plus 1.5ex minus 0ex}%
{0.7ex plus .5ex minus 0ex}{\normalfont\normalsize\itshape}}
\makeatother

%% file: supplement_refs.tex
% Generated by joint build; use starred cross-document references.
\newlabel{supp:sec:overview}{{S1}{1}{Overview}{section.1}{}}
\newlabel{supp:sec:method_supplement}{{S2}{1}{Implementation and Supervision}{section.2}{}}
\newlabel{supp:sec:s1}{{S2.A}{1}{Input Construction and Implementation}{subsection.2.1}{}}
\newlabel{supp:sec:s2}{{S2.B}{1}{Weak Semantic Supervision and Teacher Protocol}{subsection.2.2}{}}
\newlabel{supp:fig:component_supervision}{{11}{3}{Sensitivity to $\lambda _{\mathrm {sem}}$. Half violins (left) show (a--g) component scores and (h) predicted ratings; white-centered markers and black bars mark medians and interquartile ranges. Curves (right) show (a--g) weak-label Spearman correlations and (h) PCC with mean human ratings. Colors identify seeds; curves connect same-seed results}{figure.11}{}}
\newlabel{supp:sec:s3}{{S3}{3}{Extended Ablations and Sensitivity}{section.3}{}}
\newlabel{supp:sec:statistics}{{S3.A}{3}{Statistical Procedures}{subsection.3.1}{}}
\newlabel{supp:sec:s4}{{S4}{4}{Human Evaluation}{section.4}{}}
\newlabel{supp:tab:auxiliary_controls}{{IV}{5}{Auxiliary-design ablations. Intervals follow Section~\ref {sec:statistics}. Paired diagnostic values list the variant before DisFace3DNet}{table.4}{}}
\newlabel{supp:sec:s5}{{S5}{6}{Prediction Comparisons with FPEM}{section.5}{}}
\newlabel{supp:sec:s6}{{S6}{6}{Applications}{section.6}{}}